\documentclass[%
 reprint,
 amsmath,amssymb,
 aps,
]{revtex4-2}

\usepackage{graphicx}
\usepackage{dcolumn}
\usepackage{bm}
\usepackage{color}

\definecolor{my_green}{rgb}{0.55, 0.71, 0.0}

\begin{document}

\preprint{APS/123-QED}

\title{Resonant-Tunnelling Diode Reservoir Computing System for Image Recognition}

\author{A.~H.~Abbas$^*$, Hend Abdel-Ghani, and Ivan S.~Maksymov}
\affiliation{%
 Artificial Intelligence and Cyber Futures Institute,\\
 Charles Sturt University, Bathurst, NSW 2795, Australia
}

\date{\today}

\begin{abstract}
As artificial intelligence continues to push into real-time, edge-based and resource-constrained environments, there is an urgent need for novel, hardware-efficient computational models. In this study, we present and validate a neuromorphic computing architecture based on resonant-tunnelling diodes (RTDs), which exhibit the nonlinear characteristics ideal for physical reservoir computing (RC). We theoretically formulate and numerically implement an RTD-based RC system and demonstrate its effectiveness on two image recognition benchmarks: handwritten digit classification and object recognition using the Fruit~360 dataset. Our results show that this circuit-level architecture delivers promising performance while adhering to the principles of next-generation RC, eliminating random connectivity in favour of a deterministic nonlinear transformation of input signals.

\begin{description}
\item[Keywords]
Tunnel diodes, image recognition, reservoir computing, nonlinear dynamics, activation function.

\end{description}
\noindent $^*$Corresponding author: \texttt{aborae@csu.edu.au}

\end{abstract}

\maketitle
\section{Introduction}
Machine learning-based image recognition has become a cornerstone technology in scientific research \cite{Kre22}, healthcare \cite{Naj23}, defence \cite{Sua23}, agriculture \cite{Dha23} and industry \cite{Sua23, Zha24}. By enabling computers to identify and classify visual data with high accuracy, these systems are transforming how we interpret complex patterns from diagnosing medical conditions in radiology \cite{Naj23} to guiding autonomous vehicles \cite{Qiu24} and enhancing satellite imagery analysis \cite{Rol21}.

At its core, this technology learns from large datasets of images, gradually improving its ability to detect subtle features that might be missed by the human eye. Typically, an image recognition system consists of a scanner that converts real-world images and objects into digital form and a machine vision subsystem that learns from datasets of these digital images to infer the identity of previously unseen scanned inputs \cite{Tch20}.

The recognition part of such a system is mostly delegated to software running on a digital computer. Typically, the computer must have considerable computational power to store and process information. While using a high-performance workstation is not an issue in a stationary context, such as a research laboratory server room or data centre, this solution is not suitable for autonomous systems, especially airborne drones, where power supply capacity, weight and cost are critical constraints \cite{Kat24, Xia25}.

While it is technically possible to replace the software and digital computer with electronic \cite{Gra89, Yan19, Zha24} or optical \cite{Fu24} neural networks, a more viable solution may lie in a hybrid approach. In this method, certain functions of the neural network, such as the nodes and their associated activation functions, are replaced by electronic components or circuits \cite{Akt25}. This neuromorphic approach, which uses brain-inspired principles to design circuits capable of performing computational tasks with significantly greater power efficiency than conventional computers \cite{Sch22, Zha24_1, Kud25}, would still rely on a reasonably fast and low-power digital computer for data processing. However, it would be more efficient overall due to the implementation of neurons as electronic circuits \cite{Zha24_1, Kud25}.

This approach is particularly well suited to physical (hardware) reservoir computing (RC), a branch of neuromorphic computing that employs the intrinsic dynamics of physical systems to perform efficient computation \cite{Tan19, Nak20, Mak23_review, Abb24}. In particular, RC systems have proven effective for processing highly nonlinear and chaotic time-series datasets. A key advantage of RC is its low training cost, as it typically requires minimal training data and relies only on linear optimisation of the readout layer \cite{Luk09}. However, conventional RC models depend on randomly initialised weight matrices to define neuron connectivity and input mapping, which often requires careful tuning. To address this challenge, recent research has shown a mathematical equivalence between RC and nonlinear vector auto-regression (NVAR), a formulation that eliminates the need for random matrices, reduces hyperparameter complexity and offers improved interpretability \cite{Gau21, Bol21}. This development is referred to as next-generation reservoir computing.

RC systems can also be implemented using a diverse range of physical platforms, including electronic \cite{Lia24, Abb25}, optical \cite{Che20_review} and mechanical systems \cite{Abb24_1}. Among these, electronic systems are particularly promising, as they can be realised in mixed-signal or fully analogue form \cite{Yan24, Lia24}. Moreover, large-scale implementations are feasible through printed electronics, enabling the physical circuitry to emulate the structure and connectivity of software-based neural networks \cite{Gar22}.

In this paper, we investigate a novel hardware-based implementation of an electronics-based physical RC system as a step toward practical neuromorphic computing. The proposed system employs the intrinsic physical nonlinearity of tunnel diodes---specifically of RTDs since they have been identified as the promising candidates for applications in neuromorphic computing~\cite{OPiwonka21, OPiwonka21_1, Rom23}---to realise the reservoir layer, replacing the need for computationally-intensive software-defined transformations or input masking typically used in conventional algorithmic RC frameworks.

\begin{figure}[t]
    \centering
    \includegraphics[width=\linewidth]{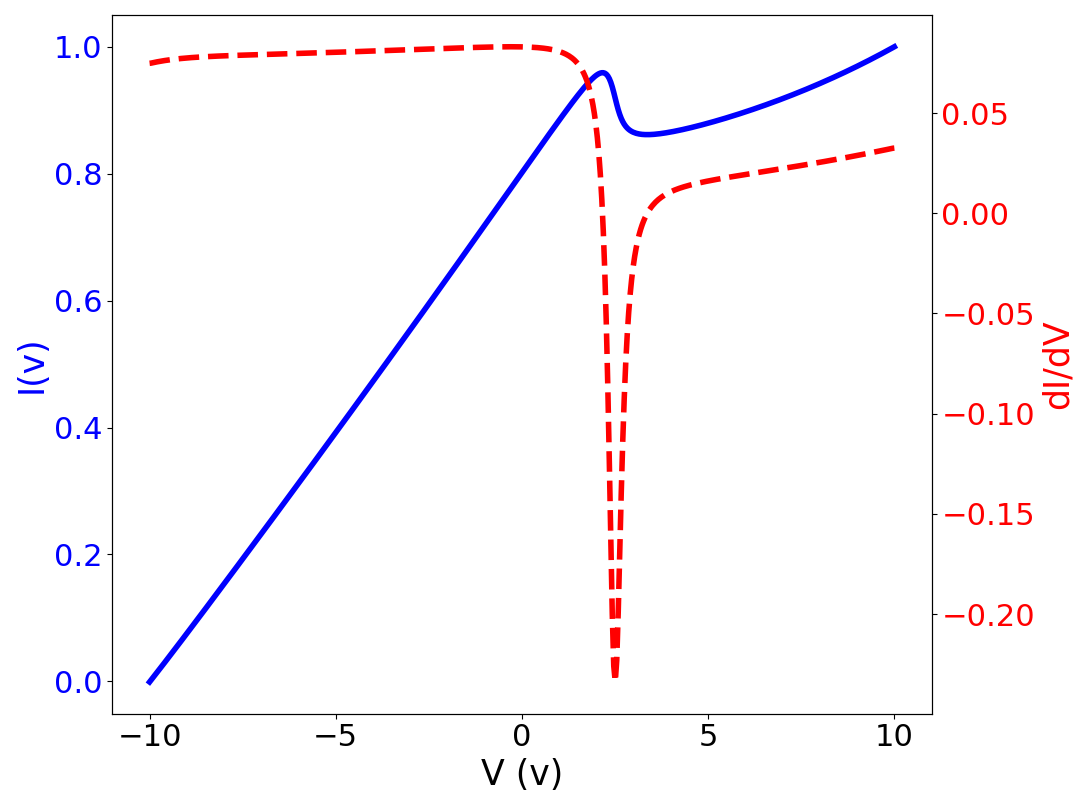}
    \caption{$I–V$ characteristic (the blue curve) and NDR (the red dashed curve) of the model RTD employed in this study.}
    \label{1}
\end{figure}

\begin{figure}[t]
\includegraphics[width=\linewidth]{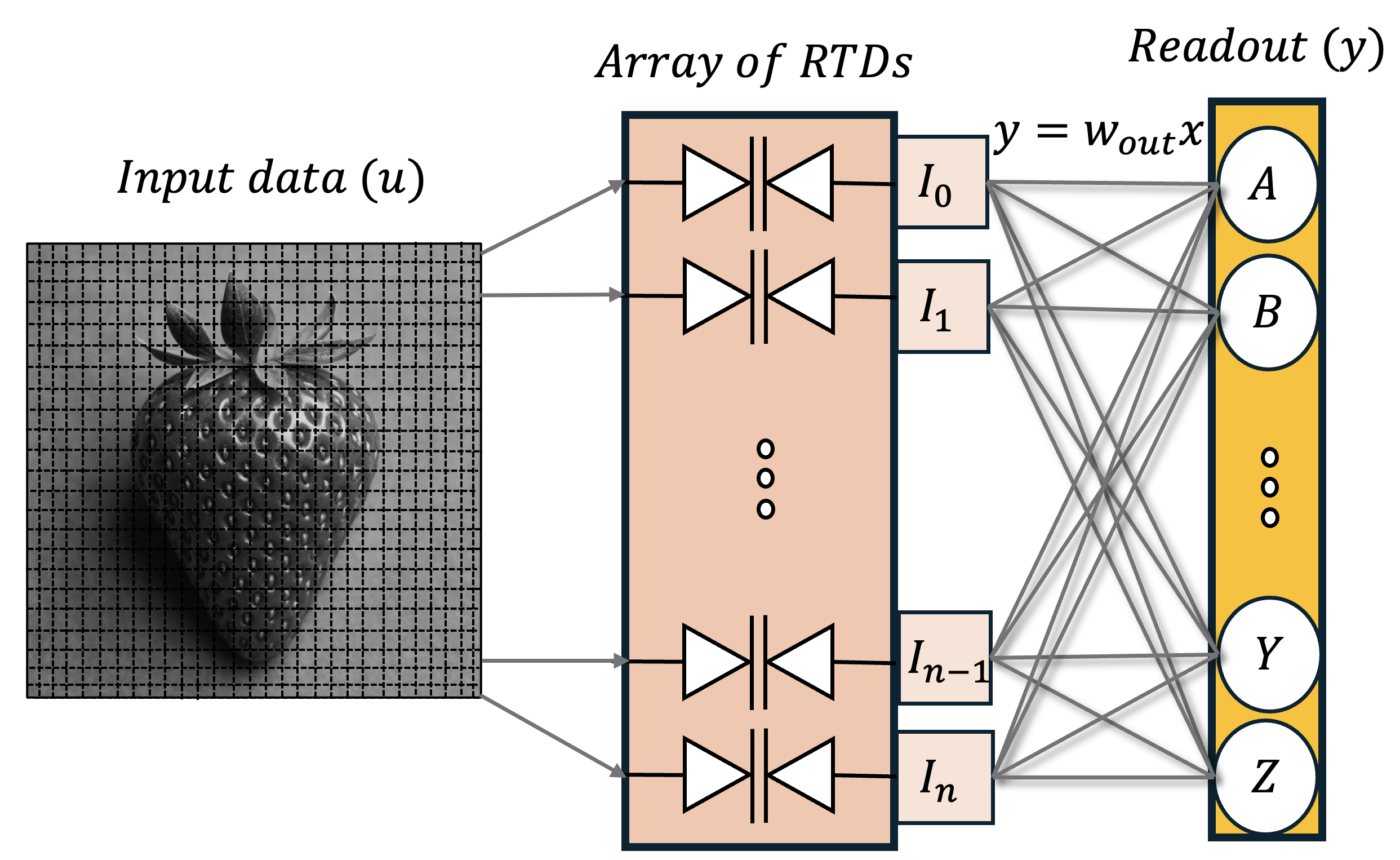}
\caption{\label{2} Schematic illustration of the proposed RTD-based RC architecture. An array of RTDs serves as a computational reservoir that facilitates, by virtue of the nonlinear $I–V$ characteristics of the diodes, the transformation of low-dimensional input signals into high-dimensional representations. The reservoir output (the currents $I_i$, $i = 0 \dots n$, produced by each diode) is processed using a linear readout mechanism.}
\end{figure}

Specifically, we explore the image recognition performance of a physical RC architecture, where an array of parallel RTDs exhibits nonlinear current–voltage ($I–V$) behaviour (Fig.~\ref{1}). This nonlinear response naturally induces the rich spatiotemporal dynamics required for high-dimensional input projection \cite{Luk09}. The system operates in a purely feedforward configuration (Fig.~\ref{2}), using the inherent $I–V$ characteristics of the diodes to perform dynamic state mapping, thus enabling efficient analogue computation without recurrent feedback or complex signal preprocessing. We also highlight that the RC architecture developed in this paper serves as a hardware counterpart to the next-generation RC algorithm \cite{Gau21}, eliminating the need for random matrices and the computationally expensive operations associated with them in traditional RC models \cite{Luk09}. The computational performance of the proposed RTD-based RC system improves as we controllably increase the nonlinearity of the response of individual RTDs.

\section{Resonant-Tunnelling Diode as a Computational Reservoir}
A tunnel (Esaki) diode is a type of heavily doped $p$–$n$ junction semiconductor device whose $I$–$V$ characteristic exhibits a region of negative differential resistance (NDR) due to quantum mechanical tunnelling~\cite{Esa58, Sze_book}. While resonant-tunnelling diodes (RTDs) differ in structure---instead of using a heavily doped junction they typically involve quantum well structures with double barriers---they share key physical principles with Esaki diodes, most notably quantum tunnelling and the resulting nonlinear transport phenomena \cite{Sze_book, Cha74, Sun98, Yan08}. Therefore, while this study employs a theoretical model developed specifically for RTDs \cite{OPiwonka21, OPiwonka21_1}, the underlying discussion is, in principle, applicable to other classes of quantum electron devices that explore the effect of quantum tunnelling. Indeed, what is important for the mainstream discussion in this present work is the fact that a voltage-dependent tunnelling behaviour of the diode gives rise to highly nonlinear responses, where the diode's resistance varies dynamically with the applied signal~\cite{Abb24_1}.

It is also instructive to comment on the suitability of RTDs as elements of an RC system. While the algorithmic RC architecture and its modifications are well established and understood \cite{Luk09}, the debate over the suitability of different physical systems for RC purposes remains ongoing, primarily because virtually any nonlinear dynamical physical system can, in principle, serve as a reservoir, although not all such systems yield efficient computational performance \cite{Ste24, Vru24, Nak24}.

Apart from their suitability for broader applications in neuromorphic computing \cite{OPiwonka21, OPiwonka21_1, Rom23}, RTDs (more specifically, their $I–V$ characteristics) can, in principle, mathematically substitute standard activation functions, potentially yielding improved overall model performance \cite{McNaughton25, Abb25_HQ}. (The same holds true for the quantum tunnelling effect model, which is relevant to the operation of an RTD \cite{Maks25_2}.) However, in the computational scheme proposed in this work, RTDs substitute the computationally demanding, nonlinear signal processing component of the reservoir computing algorithm with the physical---albeit numerically simulated in this study---response of RTDs (a hardware implementation of an array of RTDs is technologically viable \cite{Kem01, Yan08}). As a result, the RTD-based RC system promises to be more energy-efficient while delivering robust computational performance, thereby fulfilling one of the main goals of physical reservoir computing \cite{Tan19, Nak20, Mak23_review, Abb24_1}.

The $I$–$V$ characteristic of an RTD can be approximated by the expression \cite{OPiwonka21}
\begin{equation} \label{eq:1}
    I(V) = I_{t}(V)+I_{r}(V)\,,
\end{equation}
where the individual components are defined as
\begin{eqnarray} \label{eq:term1}
    I_{t}(V) &=& a\ln \left( \frac{1 + e^{\alpha + \eta V}}{1 + e^{\alpha-\eta V}} \right)  \times \left(\frac{\pi}{2} + \tan^{-1} \left( \frac{c - n_1V}{d} \right)\right)\,, \nonumber\\
    I_{r}(V) &=& h \left(e^{\gamma V} - 1 \right)\,.
\end{eqnarray}
The parameters are given by $\alpha = \dfrac{q (b - c)}{k_B T}$, $\eta = \dfrac{q n_1}{k_B T}$, and $\gamma = \dfrac{q n_2}{k_B T}$, where $q$ denotes the electron charge, $k_B$ is Boltzmann’s constant, $T$ is the temperature, $V$ is the voltage, and $a$, $b$, $c$, $d$, $n_1$, and $n_2$ are system-specific parameters. In this study, we adopt the values $T = 300$\,K, $a = 0.0039$\,A, $b = 0.5$\,V, $c = 0.0874$\,V, $d = 0.0073$\,V, $n_{1} = 0.0352$, $n_{2} = 0.0031$, and $h = 0.0367$\,A~\cite{OPiwonka21}. Unlike in Ref.~\cite{OPiwonka21}, we increase the parameter $b$ by a factor of 10 to investigate the effect of enhanced diode nonlinearity on the computational performance of the RC system under study. This amplification emulates the introduction of electronic gain into the system. Further relevant details can be found in Ref.~\cite{McNaughton25}.

The $I–V$ characteristic of the RTD calculated using Eq.~(\ref{eq:1}) has three distinct regions: two segments where the differential resistance is positive, separated by a central region exhibiting NDR (approximately within the interval \( V \in [1, 3] \) in Fig.~\ref{1}). The term \( I_{t}(V) \) in Eq.~(\ref{eq:1}) governs the low-voltage behaviour, producing both the initial positive differential resistance and the NDR region. However, this term does not capture the resurgence of current at higher voltages. In turn, the term \( I_{r}(V) \), dominant at elevated voltages, accounts for this second region of positive differential resistance~\cite{OPiwonka21}.

\section{Reservoir Computing Model}
Figure ~\ref{2} presents a schematic diagram of the proposed RC system architecture. At its core, the nonlinear $I–V$ characteristic of RTDs is exploited as the primary nonlinear element. A parallel array of RTDs constitutes the reservoir layer, with each diode contributing to the nonlinear transformation of the input signals, as explained below.

The RTD-based RC system processes grayscale images by converting them into electric voltage signals. To this end, each input image is first transformed into a binary format and then normalised to align with the diode’s operating voltage range, $V \in [-5, 5]$. Such an image processing is conducted row-wise: each horizontal row of pixels in the image is encoded into a sequence of voltage pulses applied to a corresponding RTD in the array. Thus, for an image of size $n \times m$ pixels, the reservoir comprises $n$ RTDs, each receiving a pulse train of length $m$ that represents the pixel intensities of its row. For example, with standard MNIST images of $28 \times 28$ pixels, the system utilises 28 diodes, each driven by a 28-element voltage pulse sequence. The values $I_i$, $i=0\dots n$, of the output current produced by each diode are treated as the dynamic reservoir states \cite{Luk09}, which form a high-dimensional embedding of the input and used to train a linear readout (Fig.~\ref{2}).

Furthermore, each image is associated with a one-hot encoded label vector, where the elements corresponding to the true class is set to 1 and all others to 0. For example, the digit `1' in the MNIST dataset and the fruit `Apricot' in the Fruit-360 dataset both correspond to class index 1 and are encoded as $[0, 1, 0, \dots, 0]^T$.

The image recognition procedure begins with the aforementioned pixel-to-voltage encoding, followed by the application of these voltage sequences to the RTD array. For each input image, the dynamic responses from the diodes are assembled into a reservoir state matrix $\mathbf{x} \in \mathbb{R}^{k \times j}$, where $k$ is the number of samples and $j$ is the dimensionality of the reservoir output. The corresponding label vectors form the output matrix $\mathbf{y} \in \mathbb{R}^{k \times s}$, where $s$ is the number of classes.

As the next step, the output weights are computed by applying ridge regression to minimise the mean squared error between predicted and target outputs as \cite{Luk09}
\begin{equation}
\mathbf{w}_{\text{out}} = (\mathbf{x}^\top \mathbf{x} + \lambda \mathbf{I})^{-1} \mathbf{x}^\top \mathbf{y}\,,
\end{equation}
where $\lambda$ is the regularisation parameter and $\mathbf{I}$ is the identity matrix.

As a part of the testing procedure, the reservoir states for unseen inputs, $\mathbf{x}_{\text{test}}$, are processed through the trained readout layer as
\begin{equation}
\mathbf{y}_{\text{pred}} = \mathbf{x}_{\text{test}} \mathbf{w}_{\text{out}}\,.
\end{equation}
The classification is done by selecting the index corresponding to the maximum value in the predicted output vector $\hat{\mathbf{y}}$.

The proposed RTD-based RC architecture is advantageous compared with the traditional RC algorithm \cite{Luk09}. For example, it eliminates the need for mask matrices, delay lines, virtual nodes and feedback terms that are often present in standard RC system implementations \cite{Tan19}. The only tuning required by the RTD-based RC system is the proper configuration of the pixel-to-voltage mapping function, aim to ensure that sufficient nonlinearity is present in the system's response.

\section{Results and Discussion}
To theoretically evaluate the performance of the RTD-based RC system, the equations of the RTD model and the equations of the RC model presented above were implemented as a software package in Python~3. The resulting software was tested on two benchmark image recognition datasets: the standard MNIST and Fruit-360 \cite{Mur18}. The well-known MNIST dataset contains 60,000 grayscale images of handwritten digits from 0 to 9, each ones sized $28 \times 28$ pixels. The Fruit-360 dataset includes 10,000 grayscale images representing 15 fruit classes, resized to $28 \times 28$ pixels. Both datasets were split into 70\% training and 30\% testing subsets.

All numerical experiments were conducted using a 64-bit workstation computer powered by an Apple M3 Max CPU and equipped with 128\,GB of unified memory. The input images were shuffled prior to training and testing to ensure unbiased results. The voltage mapping range applied to the RTD array was optimised and set to $[-5, 5]$\,V.
\begin{figure}[t]
    \includegraphics[width=\columnwidth]{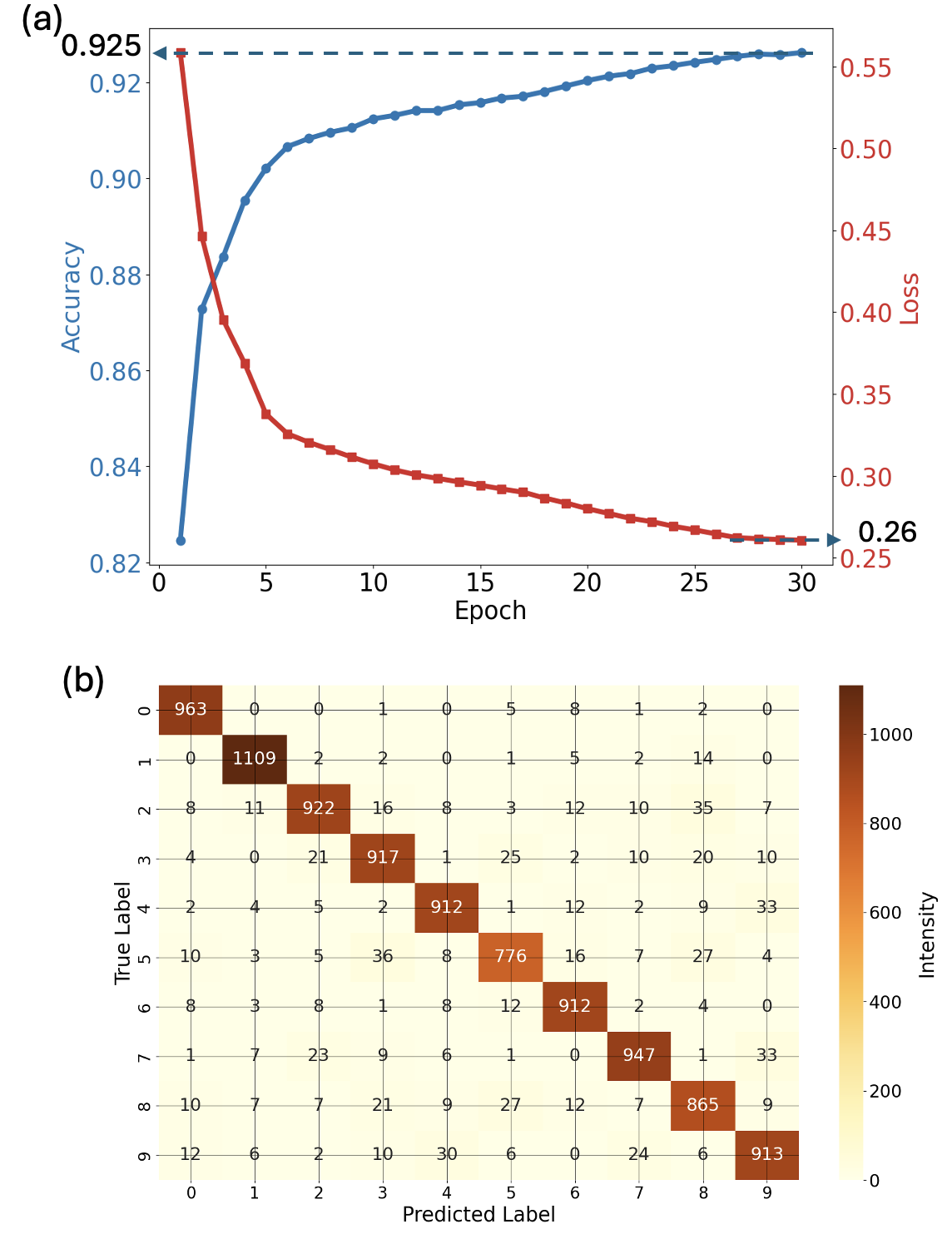}
    \caption{\label{3} Evaluation of the digit recognition model using the RTD-based RC system trained on the MNIST dataset:~(a) training performance over epochs, showing accuracy and loss, and (b) the confusion matrix, illustrating classification accuracy across digit classes.}
\end{figure}

Figure~\ref{3}(a) shows the training accuracy and training loss as functions of the number of epochs for the RTD reservoir trained on the MNIST dataset. The model is observed to converge after 30 epochs, achieving an accuracy of 92.5\% and a loss of 0.26, indicating effective learning. Figure~\ref{3}(b) shows the confusion matrix---which illustrates the distribution of predicted versus actual classifications and highlights the performance of the model across individual digit classes---where the majority of samples are correctly classified, particularly for digits `1', `0' and `7'. We can observe that misclassifications are minimal and primarily occur between visually similar digits, such as `4' and `9', suggesting the potential benefit of more sophisticated feature extraction or improved temporal encoding (see, e.g., Ref.~\cite{Maks25_1, Maks25_2} for further relevant discussion).

\begin{figure}[t]  
    \includegraphics[width=\columnwidth]{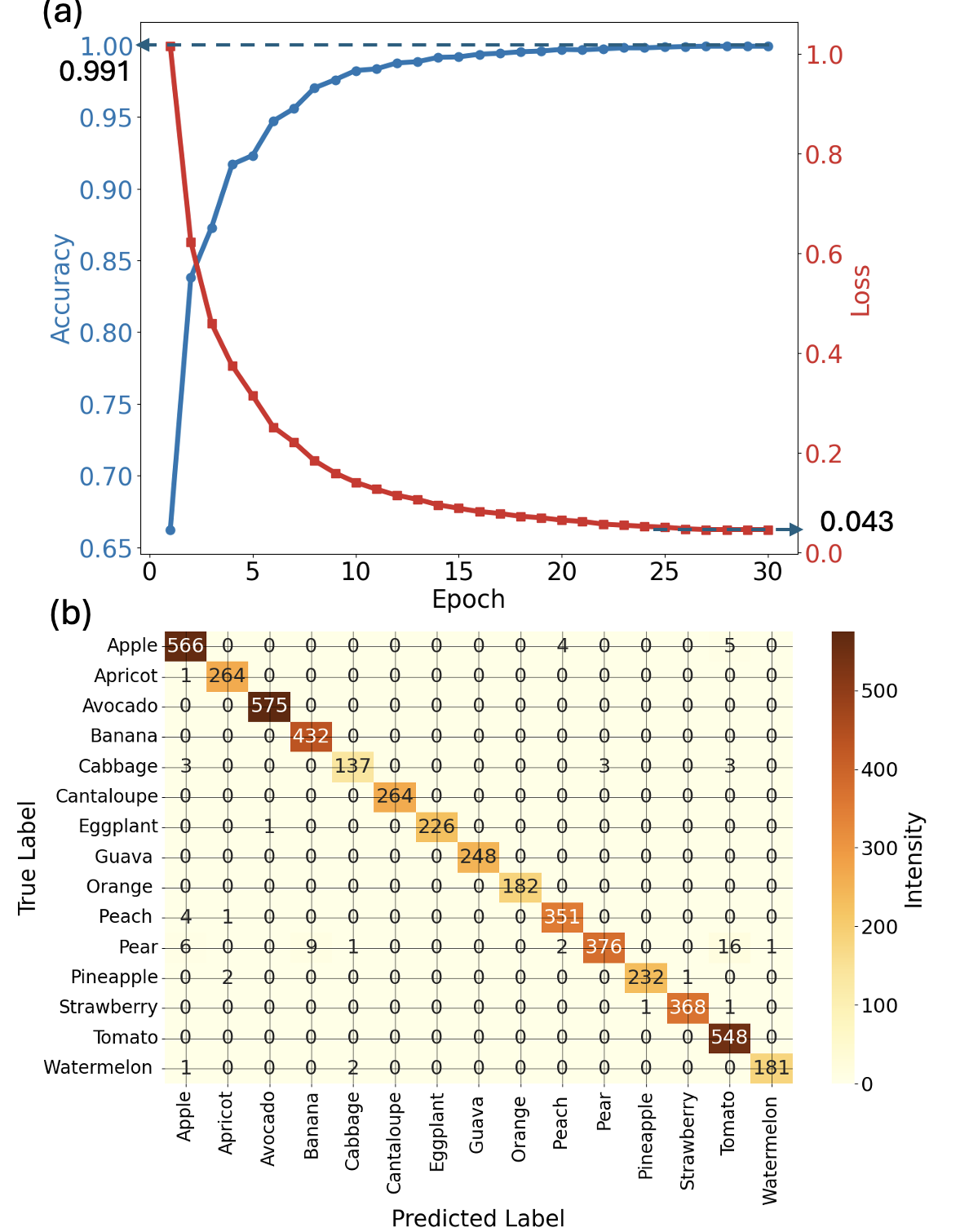}  
    \caption{\label{4} Evaluation of the digit recognition model using the RTD-based RC system trained on the Fruit-360 dataset:~(a) training performance over epochs, showing accuracy and loss, and (b) the confusion matrix, illustrating classification accuracy across fruit classes.}
\end{figure}

Figure~\ref{4} summarises the performance of the RTD-based RC system on the Fruit-360 dataset, which contains visually distinct classes of fruits. The RC system achieves a final accuracy of 99.1\% after 30 training epochs. As shown in Figure~\ref{4}(a), the accuracy increases rapidly while the loss declines consistently to a final value of approximately 0.046, indicating fast convergence and minimal overfitting \cite{Ali24}.

The confusion matrix in Fig.~\ref{4}(b) shows near-perfect classification across all 15 fruit categories. Notably, the classes `Apple', `Avocado' and `Strawberry' exhibit perfect or near-perfect accuracy, showcasing the strong discriminative ability of the model in classifying natural images with high inter-class variability.

The superior performance of the RC system on the Fruit-360 dataset compared to the MNIST handwritten digits dataset can be attributed to the greater size, complexity and variability of the images in the Fruit-360 dataset. Indeed, Fruit-360 comprises thousands of images spanning a diverse set of fruit classes, each with unique textures and shapes. This richness in visual features enables the RC system to exploit its computational ability more effectively. In contrast, MNIST contain a large number of images of only ten digit classes, with comparatively low intra-class variability and limited feature complexity. As a result, the ability of the RC system to do nonlinear transformations and temporal encoding may be underutilised on MNIST.
\begin{figure}[t]
    \includegraphics[width=\columnwidth]{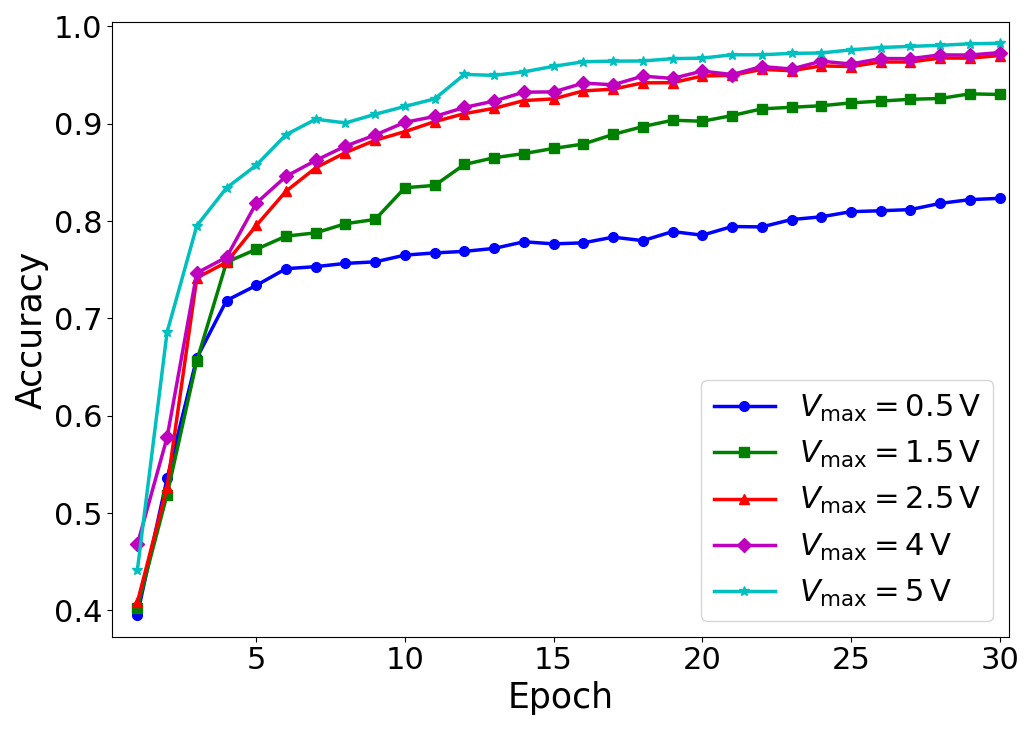}
    \caption{Training accuracy as a function of epochs for increasing values of $V_{\text{max}}$. Raising $V_{\text{max}}$ results in an increase in the strength of nonlinearity, thereby improving classification accuracy.}
    \label{5}
\end{figure}

As a next step, to evaluate the impact of the nonlinearity of the response of RTDs on the training dynamics of the RC system, numerical experiments were conducted by setting different voltage limits $V_{\text{max}}$ ranging from 0.5\,V to 5\,V. Figure~\ref{5} illustrates how varying $V_{\text{max}}$ influences the training accuracy. The results clearly show that a higher $V_{\text{max}}$ leads to better accuracy. In particular, using lower values of $V_{\text{max}}$ (e.g., 0.5\,V) restricts the input voltage range, resulting in the reservoir operating mostly within the linear region of the $I–V$ characteristic. This limits the dynamic transformation capability of the reservoir and, therefore, yields poorer accuracy of approximately 80\% (an explanation of the operation of a reservoir using a linear activation function can be found in Ref.~\cite{Jae01}). However, as $V_{\text{max}}$ increases to 4--5\,V, the reservoir accesses a richer nonlinear operating regime, enabling more expressive mappings and more effective temporal feature encoding \cite{Jae01, Luk09}. As shown in Fig.~\ref{5}, this significantly enhances performance, with final accuracies approaching or exceeding 95\%, which further confirms the critical role of nonlinearity in the reservoir’s computational power.
\begin{figure}[t]
    \includegraphics[width=\columnwidth]{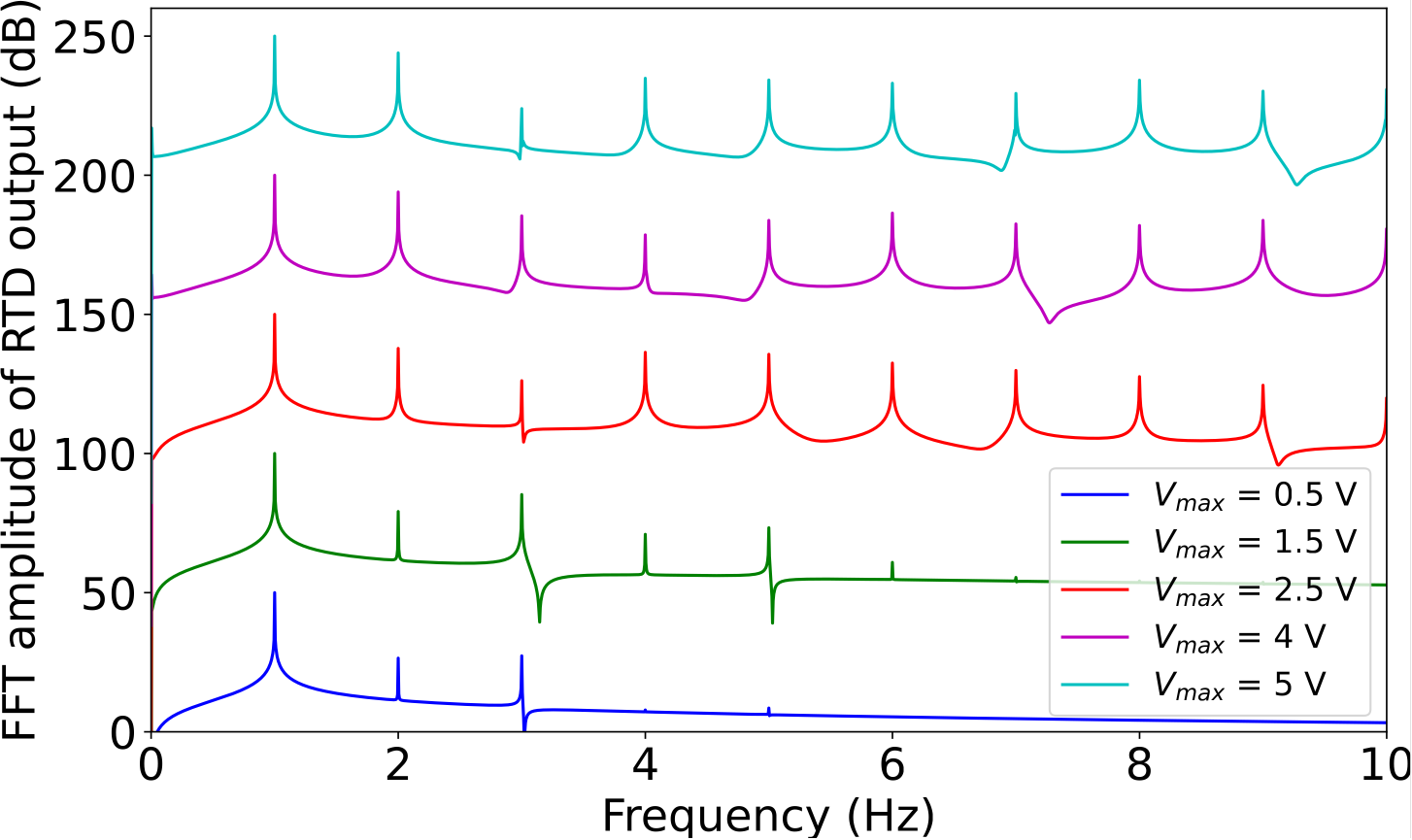}
    \caption{Fourier transform spectra of the outputs of the RTD excited by a test input sinusoidal signal. Compared with the single-frequency spectrum of the input sinusoidal signal, the RTD output is enriched by higher-order harmonics generated through nonlinear processes. The number of harmonics, and thus the strength of the nonlinear response, increases with the amplitude of the input signal, as controlled by $V_{\text{max}}$. For illustration, each spectrum is vertically shifted by 50\,dB.}
    \label{rtd_fft}
\end{figure}

\section{Connection to the Next-Generation Reservoir Computing Framework}
The observed increase in accuracy with increasing $V_{\text{max}}$ can be understood in light of the following comparison between the operation of the RTD-based RC and the principles underlying the next-generation RC framework \cite{Gau21} discussed in at the beginning of this paper.

Recall that a traditional algorithmic RC system \cite{Jae01, Luk09} uses a reservoir composed of randomly connected nodes. In Ref.~\cite{Gau21}, an alternative RC algorithm was introduced in which the state matrix ${\bf x}$ of the traditional RC system is replaced by a matrix ${\bf x}^{\text{future}}$, composed of future states ${\bf x}_n^{\text{future}}$ corresponding to the current and time-delayed discrete input data points ${\bf u}_n$ and their nonlinear functionals. This newly constructed state matrix is then used to calculate ${\bf w}^{\text{out}}$, thereby avoiding certain computationally intensive steps associated with traditional RC systems.

Furthermore, based on the results presented in Ref.~\cite{Bol21}, it has been demonstrated that this computational scheme not only circumvents the use of randomly generated neural connection matrices but is also equivalent to, and in some cases more efficient than, the traditional RC algorithm. This approach is referred to as next-generation reservoir computing \cite{Gau21, Wan25}. 

The nonlinear component of the future states ${\bf x}_n^{\text{future}}$ can be defined as an arbitrary nonlinear function of the input signal. In Ref.~\cite{Gau21}, accurate forecasts were achieved using polynomial functions, with the key insight that only a few low-order polynomial terms were sufficient to obtain reliable results. A similar finding was reported in Ref.~\cite{Gov22_1} within the context of quantum RC systems, where the authors applied a nonlinear transformation to the input data and demonstrated that subsequent processing steps typical of traditional RC algorithms were no longer necessary.

The nonlinear transformation of the input data implemented in the RTD-based RC system proposed in this work can be illustrated by applying a sinusoidal wave to one of the diodes in the RTD array shown in Fig.~\ref{2}. (Such an approach was approbated in the previous works Ref.~\cite{Tro20, Mak24_dynamics}; the choice of the frequency of the sinusoidal wave is immaterial in this scenario.) We then compute the Fourier spectra of both input sinusoidal wave and output signal produced the diode, and present them in Fig.~\ref{rtd_fft}.

The spectrum of the RTD output reveals not only the fundamental frequency peak at 1\,Hz but also additional peaks corresponding to the second, third, fourth and so on higher-order harmonics. We can see that the number of harmonics increases as the control voltage amplitude $V_{\text{max}}$ is increased. In contrast, the spectrum of the input sinusoidal wave contains only a single peak at 1\,Hz. This result demonstrates that the RTD effectively performs a nonlinear transformation of the input, effectively representing it as a polynomial-like function and thereby enriching the reservoir's feature space in a next generation RC system fashion.

This result helps explain the differences between the accuracy-vs-epochs curves shown in Fig.~\ref{5}. The most pronounced increase in accuracy is observed when $V_{\text{max}}$ increases from 0.5\,V to 1.5\,V. Spectral analysis reveals that this change results in the emergence of significant fourth and fifth nonlinear harmonics in the output spectrum in Fig.~\ref{rtd_fft}. (The colour markings of the lines in Fig.~\ref{5} and Fig.~\ref{rtd_fft} are consistent, allowing for a direct comparison between the classification accuracy trends and the spectral content of the RTD outputs at corresponding values of $V_{\text{max}}$.) A similar trend occurs when $V_{\text{max}}$ reaches 2.5\,V, where the fourth, fifth and sixth harmonics become prominent and comparable in amplitude to the second and third harmonics.

This observation highlights the advantage of retaining multiple nonlinear terms in the next generation RC framework. However, while retaining additional nonlinear terms beyond quadratic and cubic typically incurs increased computational cost in algorithmic implementations of the next-generation scheme, the RTD-based approach allows up to ten nonlinear terms to be retained simply by increasing $V_{\text{max}}$, with only a modest rise in the system’s overall energy consumption.

\section{Conclusions}
In this work, we have explored the potential of resonant-tunnelling diodes as nonlinear physical elements for reservoir computing applications. We have demonstrated that the characteristic $I–V$ nonlinearity of the diodes can effectively substitute conventional reservoir computing algorithms relying on traditional machine learning activation functions, thereby enabling efficient and robust classification of MNIST and Fruit-360 image datasets. Through numerical experiments, we have shown that resonant-tunnelling diode-based reservoirs can achieve high classification accuracy on these benchmark datasets, with performance strongly influenced by the degree of nonlinearity and dynamic range of the diode response.

We have also suggested the technological viability of hardware implementations involving arrays of resonant-tunnelling diodes, indicating their low power consumption, high-speed operation and compatibility with existing semiconductor technologies. Hence, we argue that resonant-tunnelling diodes offer a promising physical substrate for energy- and computationally efficient reservoir computing systems, especially in scenarios where compact, high-throughput and low-latency computation is desired.

Future work will focus on experimental realisation of resonant-tunnelling diode-based reservoir systems, integration with temporal encoding schemes and exploration of hybrid architectures involving quantum tunnelling effects and emerging device platforms such as spiking-neuron integrated circuits.

\bibliography{apssamp}
\end{document}